\documentclass[letterpaper, 10 pt, conference]{ieeeconf}
\IEEEoverridecommandlockouts    %
\usepackage{graphics}           
\usepackage{times}              
\usepackage{amsmath}            
\usepackage{amssymb}            
\usepackage{graphicx}
\usepackage{algorithm}
\usepackage[noend]{algpseudocode}
\usepackage{booktabs}
\usepackage{color}
\usepackage{listings}
\usepackage{subfiles}
\usepackage{hyperref}
\usepackage[nocompress]{cite} %
\definecolor{instructioncolor}{rgb}{.5,.5,.5}

\usepackage[font=small]{caption}

\def\secref#1{Sec.~\ref{#1}}
\def\figref#1{Fig.~\ref{#1}}
\def\tabref#1{Tab.~\ref{#1}}
\def\eqref#1{Eq.~(\ref{#1})}

\makeatletter
\usepackage{xspace}
\DeclareRobustCommand\onedot{\futurelet\@let@token\@onedot}
\def\@onedot{\ifx\@let@token.\else.\null\fi\xspace}
\def\eg{e.g\onedot}

\def\etal{{et~al}\onedot}
\makeatother

\def\etalcite#1{\mbox{\etal}~\cite{#1}}

\usepackage{array}
\newcolumntype{L}[1]{>{\raggedright\let\newline\\\arraybackslash\hspace{0pt}}m{#1}}
\newcolumntype{C}[1]{>{\centering\let\newline\\\arraybackslash\hspace{0pt}}m{#1}}
\newcolumntype{R}[1]{>{\raggedleft\let\newline\\\arraybackslash\hspace{0pt}}m{#1}}

\def\argmin{\mathop{\rm argmin}}

\newcommand{\RR}{\mathbb{R}}

\renewcommand{\b}[1]{\mbox{\boldmath$#1$}}

\renewcommand{\d}[1]{\b {#1}}

\newcommand{\m}[1]{{\mbox{{\sffamily\slshape{#1\/}}}}}

\newcommand{\mq}[1]{{\mbox{{\sffamily{#1}}}}}

\usepackage[acronyms, shortcuts]{glossaries}
\usepackage[normalem]{ulem}
\usepackage{multicol}
\usepackage{multirow}
\usepackage{ragged2e}
\usepackage{tabularx}
\usepackage{makecell}

\graphicspath{{pics/}}
\newcolumntype{Y}{>{\centering\arraybackslash}X}

\newcommand{\B}{\fontseries{b}\selectfont}
\newcommand{\N}{}

\DeclareRobustCommand{\rchi}{{\mathpalette\irchi\relax}}
\newcommand{\irchi}[2]{\raisebox{\depth}{$#1\chi$}} %

\newcommand{\twodkiss}{WO\,+\,2D KISS-ICP\xspace}
\newcommand{\ours}{Kinematic-ICP\xspace}
\newcommand{\threedkiss}{WO\,+\,3D KISS-ICP\xspace}
\newcommand{\pose}{\mq{T}}
\newcommand{\se}{\mathbb{SE}}
\newcommand{\odom}{\mq{O}}
\newcommand{\pert}{\Delta \mathbf{u}}
\newcommand{\pred}{\hat{\mq{T}}}
\newcommand{\set}[1]{\mathcal{#1}}
\newcommand{\spoint}{\d{s}}
\newcommand{\tpoint}{\d{q}}
\newcommand{\model}{f(\pert)}
\newcommand{\Jkinematic}{\mathbf{J}_{\text{kinematic}}}
\newcommand{\rpe}{RPE\xspace}
\newcommand{\ate}{ATE\xspace}

\newacronym{icp}{ICP}{iterative closest point}
\newacronym{slam}{SLAM}{simultaneous localization and mapping}
\newacronym{loam}{LOAM}{lidar odometry and mapping}
\newacronym{suma}{SuMa}{Surfel-based Mapping}
\newacronym{cticp}{CT-ICP}{continuous time ICP}
\newacronym{imu}{IMU}{inertial measurement unit}
\newacronym{gps}{GPS}{global positioning system}
\newacronym{ugv}{UGV}{unmanned ground vehicle}

\newcommand{\kinematic}{Kinematic-ICP\xspace}

\newcommand{\kiss}{KISS-ICP~\cite{vizzo2023ral}\xspace}
\newcommand{\kitti}{KITTI Odometry~\cite{geiger2012cvpr}\xspace}

\title{\LARGE \bf Kinematic-ICP: Enhancing LiDAR Odometry with Kinematic Constraints for Wheeled Mobile Robots Moving on Planar Surfaces}
\author{
	\begin{tabular}{cccc}
		Tiziano Guadagnino* & Benedikt Mersch* & Ignacio Vizzo*  & Saurabh Gupta    \\
		Meher V.R. Malladi  & Luca Lobefaro    & Guillaume Doisy & Cyrill Stachniss
	\end{tabular}
	\thanks{* Authors contributed equally}%
	\thanks{Tiziano Guadagnino, Benedikt Mersch, Saurabh Gupta, Meher V.R. Malladi, Luca Lobefaro, and Cyrill Stachniss are with the Center for Robotics, University of Bonn, Germany. Ignacio Vizzo and Guillaume Doisy are with Dexory, UK. Cyrill Stachniss is additionally with the Department of Engineering Science at the University of Oxford, UK, and with the Lamarr Institute for Machine Learning and Artificial Intelligence, Germany.}%
	\thanks{This work has partially been funded
		by the Deutsche Forschungsgemeinschaft (DFG, German Research Foundation) under Germany's Excellence Strategy, EXC-2070 -- 390732324 -- PhenoRob,
		by the Deutsche Forschungsgemeinschaft (DFG, German Research Foundation) under STA~1051/5-1 within the FOR 5351~(AID4Crops),
		and
		by the German Federal Ministry of Education and Research (BMBF) in the project ``Robotics Institute Germany'', grant No.~16ME0999.
	}%
}

\begin{document}
\thispagestyle{empty}
\pagestyle{empty}
\maketitle

\begin{abstract}
	LiDAR odometry is essential for many robotics applications, including 3D mapping, navigation, and simultaneous localization and mapping. LiDAR odometry systems are usually based on some form of point cloud registration to compute the ego-motion of a mobile robot. Yet, few of today's LiDAR odometry systems consider domain-specific knowledge or the kinematic model of the mobile platform during the point cloud alignment. In this paper, we present Kinematic-ICP, a LiDAR odometry system that focuses on wheeled mobile robots equipped with a 3D LiDAR and moving on a planar surface, which is a common assumption for warehouses, offices, hospitals, etc. Our approach introduces kinematic constraints within the optimization of a traditional point-to-point iterative closest point scheme. In this way, the resulting motion follows the kinematic constraints of the platform, effectively exploiting the robot's wheel odometry and the 3D LiDAR observations. We dynamically adjust the influence of LiDAR measurements and wheel odometry in our optimization scheme, allowing the system to handle degenerate scenarios such as feature-poor corridors. We evaluate our approach on robots operating in large-scale warehouse environments, but also outdoors. The experiments show that our approach achieves top performances and is more accurate than wheel odometry and common LiDAR odometry systems. Kinematic-ICP has been recently deployed in the Dexory fleet of robots operating in warehouses worldwide at their customers' sites, showing that our method can run in the real world alongside a complete navigation stack.
\end{abstract}

\section{Introduction}
\label{sec:intro}
Accurate ego-motion estimation is crucial for any mobile robot operating in an unknown environment. Traditional LiDAR odometry pipelines estimate the pose of the robot incrementally using some variant of the~\ac*{icp} algorithm, originally introduced by Besl and McKay~\cite{besl1992pami} to register static 3D shapes. Consequently, LiDAR odometry systems based on~\ac*{icp} alignment are typically agnostic to mobile robot kinematics. This often results in an unnatural motion estimation for the platform at hand, for example, instantaneous small motions along the local z-axis for a wheeled mobile robot. At the same time, it is a common practice to assume that the odometry can drift over time, but is continuous, meaning that the pose of a mobile platform in the odometric frame always evolves in a smooth way, without discrete jumps. This is standardized in the REP-105\footnote{https://www.ros.org/reps/rep-0105.html}

\begin{figure}[t]
	\centering
	\def\svgwidth{0.95\linewidth}
	\begingroup%
  \makeatletter%
  \providecommand\color[2][]{%
    \errmessage{(Inkscape) Color is used for the text in Inkscape, but the package 'color.sty' is not loaded}%
    \renewcommand\color[2][]{}%
  }%
  \providecommand\transparent[1]{%
    \errmessage{(Inkscape) Transparency is used (non-zero) for the text in Inkscape, but the package 'transparent.sty' is not loaded}%
    \renewcommand\transparent[1]{}%
  }%
  \providecommand\rotatebox[2]{#2}%
  \newcommand*\fsize{\dimexpr\f@size pt\relax}%
  \newcommand*\lineheight[1]{\fontsize{\fsize}{#1\fsize}\selectfont}%
  \ifx\svgwidth\undefined%
    \setlength{\unitlength}{300.93533734bp}%
    \ifx\svgscale\undefined%
      \relax%
    \else%
      \setlength{\unitlength}{\unitlength * \real{\svgscale}}%
    \fi%
  \else%
    \setlength{\unitlength}{\svgwidth}%
  \fi%
  \global\let\svgwidth\undefined%
  \global\let\svgscale\undefined%
  \makeatother%
  \begin{picture}(1,1.10620598)%
    \lineheight{1}%
    \setlength\tabcolsep{0pt}%
    \put(0,0){\includegraphics[width=\unitlength,page=1]{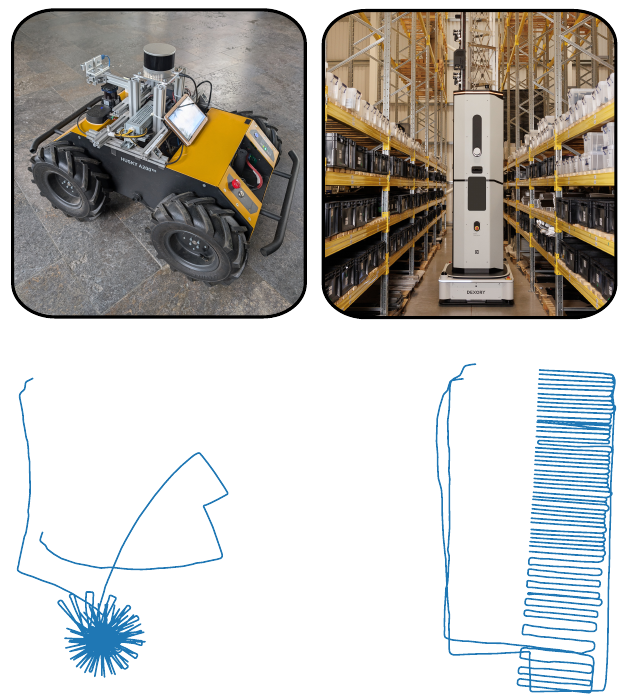}}%
    \put(0.1058124,0.38400573){\makebox(0,0)[lt]{\lineheight{1.25}\smash{\begin{tabular}[t]{l}Wheel\end{tabular}}}}%
    \put(0.40206058,0.33343942){\makebox(0,0)[lt]{\lineheight{1.25}\smash{\begin{tabular}[t]{l}\ours\end{tabular}}}}%
    \put(0,0){\includegraphics[width=\unitlength,page=2]{motivation.pdf}}%
    \put(0.53731146,0.21651241){\makebox(0,0)[lt]{\lineheight{1.25}\smash{\begin{tabular}[t]{l}$\Delta x$\end{tabular}}}}%
    \put(0.59753768,0.17864187){\makebox(0,0)[lt]{\lineheight{1.25}\smash{\begin{tabular}[t]{l}$\Delta \theta$\end{tabular}}}}%
    \put(0,0){\includegraphics[width=\unitlength,page=3]{motivation.pdf}}%
    \put(0.0776745,0.35127653){\makebox(0,0)[lt]{\lineheight{1.25}\smash{\begin{tabular}[t]{l}Odometry\end{tabular}}}}%
    \put(0,0){\includegraphics[width=\unitlength,page=4]{motivation.pdf}}%
  \end{picture}%
\endgroup%

	\caption{Given the locally consistent but globally drifting wheel odometry of a mobile robot, our approach refines the odometry using LiDAR data and a kinematic model of the platform. The depicted trajectory computed by our approach is roughly 10\,km.}
	\label{fig:motivation}
	\vspace{-0.3cm}
\end{figure}

In this paper, we introduce \kinematic, a LiDAR odometry system that explicitly incorporates robot kinematics into the~\ac*{icp} optimization. We target a motion estimation system, which is more consistent given the motion of robots. In particular, we focus on wheeled mobile robots operating on planar surfaces, as these are widely used in real-world applications, especially indoors. Our system builds upon \kiss and it uses a standard point-to-point error metric in the optimization. We use the wheel odometry of the robot as an initial guess and constrain the optimization to output a pose that follows the robot's specific kinematics. Our approach can be used on different wheeled mobile robots. Furthermore, we investigate an adaptive way to adjust the relative importance of LiDAR odometry vs. wheel odometry in the optimization to handle challenging scenarios such as feature-deficient corridors or wheel odometry inaccuracies.

The main contribution of this paper is a novel 3D LiDAR odometry system that is tightly coupled with the kinematic model of a wheeled mobile robot. Our system estimates the robot pose accurately and faster than the sensor frame rate, even in challenging scenarios where current state-of-the-art systems do not perform well. We claim that our approach can
(i) correct the wheel odometry of a mobile robot while enforcing kinematic constraints on the estimated motion;
(ii) compute the odometry with a level of accuracy on par or better than state-of-the-art LiDAR odometry systems;
(iii) can dynamically adjust the weighting between LiDAR measurements and wheel odometry readings, improving robustness and accuracy in diverse and challenging environments.
All claims are backed up by the paper and our experimental evaluation and we provide an open-source implementation at:~\url{https://github.com/PRBonn/kinematic-icp}.

\section{Related Work}
\label{sec:related}
There are many practical solutions proposed for LiDAR odometry in the context of mobile robotics~\cite{vizzo2023ral, dellenbach2022icra, deschaud2018icra, guadagnino2022ral, serafin2015iros, pan2021icra-mvls, zhang2014rss, della-corte2018icra}. Most of these methods build upon the~\ac*{icp} algorithm~\cite{besl1992pami, chen1991iros} to align consecutive LiDAR point clouds with robust kernels~\cite{chebrolu2021ral} for outlier handling to estimate the ego-motion of the sensor. Although some algorithms include proprioceptive sensors to improve the estimation~\cite{wu2024icra, wu2023ral, xu2021ral-fafr}, LiDAR odometry approaches often do not systematically consider the kinematics of wheeled mobile robots. Our work aims to change this.

A common way to introduce domain-specific knowledge into LiDAR odometry systems is to assume that the surface the robot is moving on, and thus the robot's motion, is planar. In this way, one can extract ground points in a pre-processing step of the pipeline. LeGO-LOAM~\cite{shan2018iros} computes planar and edge features and employs a two-step optimization by first estimating the platform's alignment with the ground given the planar features. The result of this step is then used as an initial guess to estimate the planar motion using the edge features. Based on this, Seo~\etalcite{seo2022ur} investigate the effect of ground segmentation and propose a robust ground-optimized LiDAR odometry system. Lately, Casado Herraez~\etalcite{casado2025icra} estimate the ground plane for an additional point-to-plane matching in radar-only odometry. Our approach focuses on wheeled mobile robots operating on a planar surface, and our estimate is purely planar. In contrast to the aforementioned methods, we do not model the planar environment based on ground segmentation, but restrict the estimated motion to follow the kinematics of the platform.

Another way to estimate the motion of ground robots is to fuse multiple sources of information in a factor graph optimization framework~\cite{grisetti2020robotics,wei2022icra, su2021ras, okawara2024arxiv, merfels2016iros, bazzana2023ral}. GR-LOAM~\cite{su2021ras} fuses LiDAR scans' feature alignment,~\ac*{imu}, and wheel encoder readings in a single factor graph, combining exteroceptive and proprioceptive information. GCLO~\cite{wei2022icra} jointly optimizes odometry and ground alignment factors to impose planarity of motion. Okawara~\etalcite{okawara2024arxiv} proposes a LiDAR-IMU system that integrates wheel odometry factors based on a kinematic model with an online calibration of the model parameters. These approaches estimate the robot's pose in a smoothing framework by considering multiple sensor cues simultaneously. We in contrast follow a different approach and include domain knowledge about the robot's motion by constraining the pose estimation from the kinematic perspective in our optimization framework.

The kinematics of mobile robots for odometry estimation have already been explored in visual odometry research. Li~\etalcite{li2022ral-viow} aims to improve the accuracy of visual inertial odometry by exploiting a velocity-control-based kinematic model without the need for odometry encoders. The motion model is calibrated online while simultaneously estimating the sensor pose using factor graph optimization. Zheng~\etalcite{zheng2019icra} parameterize the motion of a ground robot equipped with a camera in~$\se(2)$ but include the out-of-$\se(2)$ perturbations in~$\se(2)$-landmark constraints. Instead of adding soft kinematic constraints that still allow for 3D motions, other odometry approaches directly estimate the kinematic motion of a ground robot. Scaramuzza~\cite{scaramuzza2011ijcv} shows that by exploiting a restrictive Ackermann model for visual odometry, the number of correspondences for motion estimation can be reduced to one by only estimating the yaw angle around the instantaneous center of rotation. To account for modeling errors, Jordan~\etalcite{jordan2017ecmr} estimate the motion of a differential drive robot from images while falling back to full~$\se(2)$ parameterization if the model can not sufficiently explain the motion of the platform. These approaches work with camera images, whereas ours estimates the odometry from LiDAR point clouds. Furthermore, in these methods, the robot kinematics is only included to reduce the search space for data association or as an additional cue to the optimizer. Similar to Scaramuzza~\cite{scaramuzza2011ijcv}, our approach directly includes kinematics in the optimization loop, ensuring that each motion estimate is feasible for a wheeled mobile robot.

\section{LiDAR Odometry Using Point-to-Point ICP}
\label{sec:kiss}
Our approach integrates the kinematic model of a wheeled robot into the registration scheme to sequentially estimate its motion. The main components are built upon \kiss, which we briefly review in this section for completeness and to introduce the notation.

To obtain the pose~$\pose_t\,{\in}\,\se(3)$ of the robot in the odometry frame at time~$t$, we first pre-process the incoming point cloud~$\set{P}\,{=}\,\{ \d{p}_i \,{\mid}\, \d{p}_i \,{\in}\, \RR^3 \}$ expressed in the sensor frame by de-skewing and voxel downsampling resulting in~$\set{\hat{P}}^*$. We then transform this point cloud to the body frame by means of the extrinsic calibration~$\mq{C}\,{\in}\,\se(3)$ between the sensor and the body frame, resulting in a point cloud~$\set{S}\,{=}\,\{ \spoint_i \,{\mid}\, \spoint_i \,{\in}\, \RR^3 \}$. Given the previous estimate of the robot pose~$\pose_{t-1}$ and a relative odometry measurement~$\odom_{t}\,{\in}\,\se(3)$, we compute an initial guess for the current robot pose as

\begin{equation}
	\pred_{t} = \pose_{t-1}\,\odom_{t}.
\end{equation}

We then refine this estimate by using the point-to-point~\ac*{icp} algorithm. At each iteration, we obtain a set of correspondences between the source~$\set{S}$ and our local map points~$\set{Q} \,{=}\, \{ \d{q}_i \,{\mid}\, \d{q}_i \,{\in}\, \RR^3\}$ which are stored in a voxel grid as in~\kiss. We define the residual~$\mathbf{r}$ between the point~$\tpoint$ and the point~$\spoint$ transformed by~$\pose$ as
\begin{align}
	\label{eq:residual}
	\mathbf{r}(\pose) = \pose \spoint - \tpoint.
\end{align}

We then define our point-to-point cost function as:
\begin{align}
	\label{eq:cost}
	\rchi(\pred_{t}) = \frac{1}{|\set{C}|}\sum_{\left(\spoint,\tpoint\right) \in \set{C}} \left\| \mathbf{r}\left(\pred_{t}\right) \right\|^{2}_{2},
\end{align}
where~$\set{C}$ is the set of nearest neighbor correspondences, and~$|\set{C}|$ is the number of such correspondences. We can then minimize~\eqref{eq:cost} in a least squares fashion as:
\begin{align}
	\label{eq:optimization}
	\pert = \argmin_{\pert} \rchi(\pred_{t}\boxplus\pert),
\end{align}
where~$\pert$ is the~\ac*{icp} correction vector, and~$\boxplus$ applies the correction vector to the current pose estimate. This process, including nearest neighbor correspondence search and least squares optimization, is repeated until convergence, resulting in the new pose estimate~$\pose_{t}$. After convergence, we update the map with a downsampled version of the registered scan.

\section{Introducing Kinematic Constraints in ICP}
\label{sec:optimization}
Given a correction vector~$\pert\,{\in}\,\mathbb{R}^{N}$ coming from one ICP iteration, we can update our pose estimate~$\pred_{t}$ using:

\begin{equation}
	\label{eq:kinematic_update}
	\pred_{t} \boxplus \pert = \hat{\pose}_{t}\, \text{Exp}(\model),
\end{equation}
where~$f\,{:}\,\mathbb{R}^{N}\,{\rightarrow}\,\mathbb{R}^{6}$ represents the integrated kinematic model~\cite{siciliano2010robotics} of the mobile platform in use, and Exp is the exponential mapping of~$\se(3)$. The function~$f$ is used to correct the initial guess~$\pred$ coming from the platform odometry. As such, one can potentially use a kinematic model which is different than the physical configuration of the mobile platform, as in fact we are modeling a motion correction, and not the relative motion between the two scans. Intuitively,~$\pert \in \RR^{N}$ represents the~$N$ integrated control inputs that need to be applied to the robot to correct the motion.
While our framework can be applied to any robot kinematics, we want to focus on the unicycle-based correction, as this can be widely applied to most wheeled mobile robots~\cite{siciliano2010robotics}, humanoids~\cite{cognetti2017icra}, and even boats~\cite{wang2023oe}. In this case,~$\pert\,{=}\,[\Delta x, \Delta \theta]^{\top} \in \RR^2$ represent the linear and angular displacements and

\begin{equation}
	\label{eq:unicycle}
	\model      = \begin{bmatrix} \Delta x\,\frac{\sin(\Delta \theta)}{\Delta \theta + \epsilon}& \Delta x\,\frac{1 - \cos(\Delta \theta)}{\Delta \theta + \epsilon} & 0 & 0 & 0 & \Delta\theta \end{bmatrix}^{\top},
\end{equation}
where~$\epsilon$ is a small constant, typically set to the minimum positive number that can be represented as floating point.

When solving~\eqref{eq:optimization} in a least squares fashion, we need to compute the Jacobian of~\eqref{eq:residual} with respect to the correction vector~$\pert$. Looking at~\eqref{eq:kinematic_update} this Jacobian can be computed using the chain rule as:

\begin{equation}
	\begin{aligned}
		\label{eq:jacobian}
		\mathbf{J}(\pred_{t}) & =\frac{\partial \mathbf{r}(\pred_{t}\boxplus\pert)}{\partial \pert} \Big\rvert_{\pert = \mathbf{0}}                                                                                     \\
		                      & = \underbrace{\frac{\partial \pred_{t}\, \text{Exp}(\Delta \mathbf{x})\,\spoint}{\partial \Delta \mathbf{x}}\Big\rvert_{\Delta \mathbf{x} = f(\mathbf{0})}}_{\mathbf{J}_{\text{icp}}}\,
		\underbrace{\frac{\partial \model}{\partial \pert}\Big\rvert_{\pert = \mathbf{0}}}_{\mathbf{J_{\text{kinematic}}}}                                                                                              \\
		                      & =\hat{\m R}_{t}\;\begin{bmatrix} \mathbf{I} & -[\spoint]_{\times}\end{bmatrix}\;\Jkinematic,
	\end{aligned}
\end{equation}
where~$\hat{\m R}_{t}$ is the rotation part of~$\hat{\pose}_{t}$,~$\mathbf{I}\,{\in}\,\mathbb{R}^{3\times3}$ is the identity matrix, and~$[\spoint]_{\times}\,{\in}\,\mathbb{R}^{3\times3}$ is the skew symmetric matrix computed from the point~$\spoint\in\RR^{3}$. In the case of the unicycle-based correction expressed through~\eqref{eq:unicycle}, we can compute~$\Jkinematic\,{\in}\,\RR^{6\times2}$ as:
\begin{equation}
	\begin{aligned}
		\label{eq:jacob}
		\Jkinematic & =\begin{bmatrix} 1 & 0 & 0 & 0 & 0 & 0 \\ 0 & 0 & 0 & 0 & 0 & 1\end{bmatrix}^{\top},
	\end{aligned}
\end{equation}

In wheeled mobile robot systems, the rotation estimate is often noisier than the translation due to factors like wheel slippage, mechanical wear, or uneven surfaces. Wheel encoders often provide reliable translation estimates but tend to be less reliable for rotation, especially in the presence of external disturbances. In practice, we would like our LiDAR correction to focus more on the rotational part of the estimate while trusting the translational part coming from the wheel encoders more. This can help the system in degenerate scenarios in which LiDAR scans cannot completely determine the robot pose,~\eg, moving in a straight featureless corridor or entering through a narrow passage into a previously unseen part of the environment.
This knowledge can be introduced in the optimization by adding a regularization term to the cost function as:
\begin{equation}
	\label{eq:regularized_cost}
	S(\pred_{t}) = \rchi(\pred_{t}) + \frac{1}{\beta_{t}}\,\Big\|\text{Log}_{\mathbf{t}}\underbrace{\left(\pose_{t-1}\,\odom_{t}\,\pred^{-1}_{t}\right)}_{\mq{D}_t}\Big\|^{2}_{2},
\end{equation}
where $\mq{D}_t\in\se(3)$ is the deviation of the current estimate from the wheel odometry initial guess, and $\text{Log}_{\mathbf{t}}(\mq{D}_t)$ extract the translation part of $\mq{D}_t$.~\eqref{eq:regularized_cost} is a highly non-linear cost function, but we can approximate it, by elegantly regularizing the translational part of $\pert$ as:

\begin{equation}
	\label{eq:regularized_cost_simplified}
	G(\pred_{t}\boxplus\pert) = \rchi(\pred_{t}\boxplus\pert) + \frac{1}{\beta_{t}}\,\Delta x^{2},
\end{equation}
where $\Delta x\,{\in}\,\RR$ is the translational part of $\pert$, and $\beta_{t}$ is inversely proportional to the amount of regularization that we want to impose on $\Delta x$. In our case, we would like to have a large value of $\beta_{t}$ in scenarios where the wheel odometry is unreliable, and thus the system should focus more on the LiDAR measurement. Conversely, we would like to have a small value of $\beta_{t}$ when the system should trust the wheel odometry more. To achieve this behavior while avoiding manual parameter tuning, we compute the value of $\beta_{t}$ in a data-driven fashion as:
\begin{equation}
	\label{eq:adaption}
	\beta_{t} = \rchi(\pose_{t-1}\,\odom_{t}),
\end{equation}
which means we evaluate~\eqref{eq:cost} at the wheel odometry initial guess, effectively considering the consistency between the wheel odometry and the LiDAR measurements. As we demonstrate in the experiments, the introduction of $\beta_{t}$ improves the system's robustness, particularly in degenerate scenarios, while improving the accuracy of the system.

\section{Experimental Evaluation}
\label{sec:exp}
The main focus of this paper is a LiDAR odometry system that incorporates a kinematic model on the pose optimization to better estimate the motion of wheeled mobile robots. The experiments reported here support our key claims, which are that our approach can (i) correct the wheel odometry of a mobile robot while enforcing kinematic constraints of the estimated motion;
(ii) compute the odometry with a level of accuracy on par or better than state-of-the-art LiDAR odometry systems;
(iii) dynamically adjust the weighting between LiDAR measurements and wheel odometry readings, improving robustness and accuracy in diverse and challenging environments.

\subsection{Experimental Setup}
We test our approach on two different platforms and two different environments to demonstrate its effectiveness and applicability to different wheeled mobile robots.

\subsubsection{Datasets}
\label{sec:datasets}
First, we run our method on data collected in a real warehouse, by the Dexory robot, which features an extendable 12\,m tower. The robot has a differential drive with caster wheels in the front and the back. It is equipped with a 90\,°\,{$\times$}\,360\,° hemispherical 32-beam Bpearl LiDAR from Robosense that stream point clouds at 10\,Hz. It also provides odometry from the wheel encoders.
The data collection was conducted in three real-world warehouse environments of varying sizes: a small site measuring 0.35\,ha, a medium-sized site of 2.3\,ha, and a large logistics site spanning 9.45\,ha. The characteristics of the sequences are summarized in \tabref{tab:dexory-datasets}.
The data was recorded without interrupting ongoing operations, resulting in sequences containing numerous dynamic objects, including forklifts and operators. Due to the considerable size of these warehouses and operating in production, it was not feasible to install motion capture systems or other reference systems for evaluation purposes. Nevertheless, to quantitatively assess the performance of our approach, we use the industry-standard Cartographer~\cite{hess2016icra}, a widely adopted SLAM framework, which provides accurate mapping and localization in complex environments as a reference to our open-loop trajectory estimation.

\begin{figure}[t]
	\centering
	\fontsize{8}{9}\selectfont
	\def\svgwidth{0.7\linewidth}
	\begingroup%
  \makeatletter%
  \providecommand\color[2][]{%
    \errmessage{(Inkscape) Color is used for the text in Inkscape, but the package 'color.sty' is not loaded}%
    \renewcommand\color[2][]{}%
  }%
  \providecommand\transparent[1]{%
    \errmessage{(Inkscape) Transparency is used (non-zero) for the text in Inkscape, but the package 'transparent.sty' is not loaded}%
    \renewcommand\transparent[1]{}%
  }%
  \providecommand\rotatebox[2]{#2}%
  \newcommand*\fsize{\dimexpr\f@size pt\relax}%
  \newcommand*\lineheight[1]{\fontsize{\fsize}{#1\fsize}\selectfont}%
  \ifx\svgwidth\undefined%
    \setlength{\unitlength}{515.59318002bp}%
    \ifx\svgscale\undefined%
      \relax%
    \else%
      \setlength{\unitlength}{\unitlength * \real{\svgscale}}%
    \fi%
  \else%
    \setlength{\unitlength}{\svgwidth}%
  \fi%
  \global\let\svgwidth\undefined%
  \global\let\svgscale\undefined%
  \makeatother%
  \begin{picture}(1,1.11253932)%
    \lineheight{1}%
    \setlength\tabcolsep{0pt}%
    \put(0,0){\includegraphics[width=\unitlength,page=1]{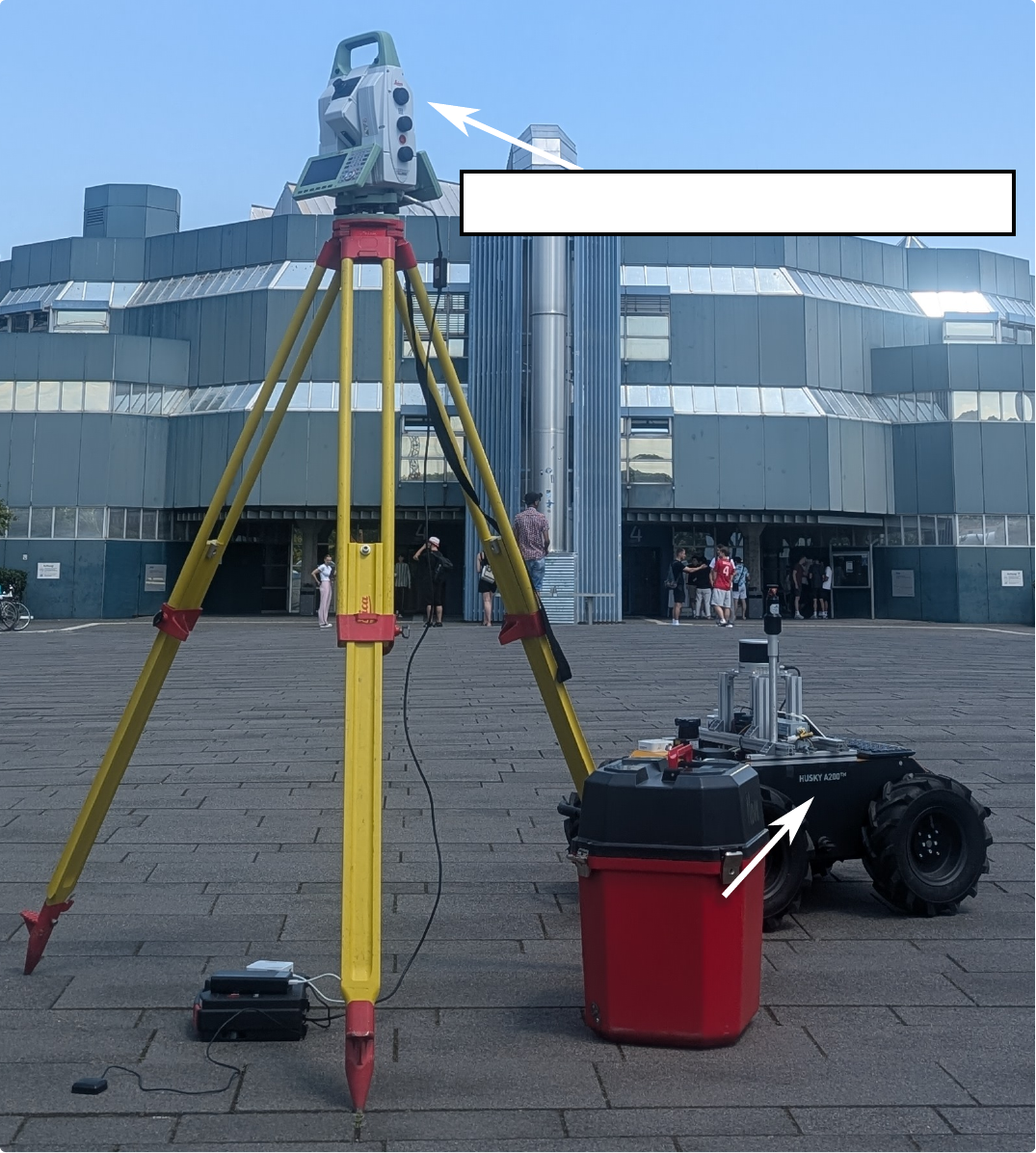}}%
    \put(0.47374541,0.89843249){\makebox(0,0)[lt]{\lineheight{1.25}\smash{\begin{tabular}[t]{l}Leica Total Station MS60\end{tabular}}}}%
    \put(0,0){\includegraphics[width=\unitlength,page=2]{reference.pdf}}%
    \put(0.52948872,0.19471617){\makebox(0,0)[lt]{\lineheight{1.25}\smash{\begin{tabular}[t]{l}Clearpath Husky\end{tabular}}}}%
    \put(0,0){\includegraphics[width=\unitlength,page=3]{reference.pdf}}%
    \put(0.33950901,0.56008622){\makebox(0,0)[lt]{\lineheight{1.25}\smash{\begin{tabular}[t]{l}Hesai LiDAR\end{tabular}}}}%
    \put(0,0){\includegraphics[width=\unitlength,page=4]{reference.pdf}}%
    \put(0.67712122,0.60391919){\makebox(0,0)[lt]{\lineheight{1.25}\smash{\begin{tabular}[t]{l}Tracking Prism\end{tabular}}}}%
  \end{picture}%
\endgroup%

	\caption{Leica total station prose tracking reference system}
	\label{fig:reference}
\end{figure}

The second platform we use to evaluate our approach is a Clearpath Husky A200, which is a four-wheeled with skid steering. The robot provides a wheel odometry estimate using its wheel encoders and is equipped with a Hesai LiDAR XT32, which records data at 10\,Hz. We drive and record data on a wide and flat pavement on our campus and in a park in front of the Poppelsdorf Palace in Bonn, which consists of uneven and rough terrain such as grass and curb walks. Note that this second location is challenging because it does not strictly follow our assumption of operating on a flat surface. To evaluate the accuracy of our approach, we mount a reflective prism on the robot and track the prism with a Leica Nova MS60 total station as shown in~\figref{fig:reference}. The total station and the robot are initially time-synchronized. The prism can be tracked with an angular accuracy of 0.0003\,$^{\circ}$ and a range accuracy of 3\,mm or better. This type of reference system has been previously used to accurately track the pose of a robot~\cite{vaidis2021crv,vaidis2023icra,vaidis2023iros}.

We point out that our system estimates the ego-motion in a robot-centric fashion, as it consider the motion model of the platform. It is essential that the extrinsic calibration between the LiDAR scanner and base frame of the robot is available when computing the odometry using Kinematic-ICP.

\begin{table}[t]
	\begin{tabularx}{\columnwidth}{c|c|c|c|c}
		\toprule
		\textbf{Sequence} & \textbf{Scans [\#]} & \textbf{Duration} & \textbf{Area}          & \textbf{Path Length} \\
		\midrule
		Small             & 13K                 & 22\,min           & 70\,m{$\times$}50\,m   & 464\,m               \\
		Mid               & 50K                 & 80\,min           & 210\,m{$\times$}110\,m & 4120\,m              \\
		Large             & 110K                & 180\,min          & 450\,m{$\times$}210\,m & 9500\,m              \\
		\bottomrule
	\end{tabularx}
	\caption{Warehouse datasets used for evaluation}
	\label{tab:dexory-datasets}
	\vspace{-0.3cm}
\end{table}

\begin{table*}[t]
	\newcommand{\seq}[1]{\multicolumn{2}{c|}{\textbf{#1}}}
	\begin{tabularx}{2\columnwidth}{c||Y|Y|Y|Y|Y|Y|Y|Y|Y|Y|Y|Y|Y|Y}
		\toprule
		                & \seq{Campus 0}   & \seq{Campus 1}   & \seq{Campus 2}   & \seq{Palace}     & \seq{Warehouse}  & \seq{Warehouse}  & \multicolumn{2}{c}{\textbf{Warehouse}}                                                                                                                                      \\

		                & \seq{}           & \seq{}           & \seq{}           & \seq{}           & \seq{Small}      & \seq{Mid}        & \multicolumn{2}{c}{\textbf{Large}}                                                                                                                                          \\
		\textbf{Method} & \rpe             & \ate             & \rpe             & \ate             & \rpe             & \ate             & \rpe                                   & \ate             & \rpe             & \ate             & \rpe             & \ate             & \rpe             & \ate             \\
		\midrule
		Wheel Odometry  & \N 4.93          & \N 3.25          & \N 4.71          & \N 6.52          & \underline{2.63} & \N 1.87          & \underline{2.98}                       & \N 4.66          & \N 2.35          & \N 1.74          & \N 0.89          & \N 16.62         & \N 4.86          & \N 108.11        \\
		\kiss           & \N 4.90          & \N 0.31          & \N 5.37          & \N 0.50          & \N 8.82          & \N 0.29          & \N 3.75                                & \underline{0.69} & \N 59.70         & \N 7.36          & \N 18.99         & \N 54.70         & \N 164.78        & \N 23.50         \\
		\threedkiss     & \N 4.64          & \N 0.34          & \underline{3.82} & \underline{0.32} & \N 5.97          & \N 0.31          & \N 4.14                                & \B 0.29          & \N 5.32          & \N 5.40          & \N 5.85          & \N 28.71         & \N 1.48          & \N 17.52         \\
		\twodkiss       & \N 4.43          & \N 0.32          & \N 3.94          & \B 0.26          & \N 6.77          & \underline{0.25} & \N 4.63                                & \B 0.29          & \N 2.02          & \N 1.01          & \N 1.01          & \underline{6.48} & \N 1.13          & \B 4.11          \\
		EKF             & \N 6.28          & \N 0.47          & \N 5.84          & \N 1.61          & \N 6.13          & \N 0.38          & \N 4.98                                & \N 0.78          & \N 0.87          & \N 0.86          & \N 0.90          & \N 9.53          & \underline{0.85} & \N 13.85         \\
		Fuse            & \underline{4.16} & \underline{0.30} & \N 6.69          & \N 0.85          & \N 3.92          & \N 0.30          & \N 3.23                                & \underline{0.69} & \underline{0.61} & \underline{0.40} & \underline{0.69} & \N 9.18          & \N 1.36          & \N 4.99          \\
		\ours           & \B 2.97          & \B 0.28          & \B 2.93          & \N 0.42          & \B 2.13          & \B 0.22          & \B 2.38                                & \N 1.56          & \B 0.53          & \B 0.26          & \B 0.46          & \B 4.44          & \B 0.39          & \underline{4.42} \\
		\bottomrule
	\end{tabularx}
	\caption{Quantitative of our approach in different environments. We report the \rpe in [\%] and the \ate in [m]. The best results are in bold, and the second best are underlined.}
	\label{tab:quantitative}
\end{table*}

\subsubsection{Baselines}
\label{sec:baselines}
We evaluate our method against the original version of \kiss and with two variants that incorporate the robot's wheel odometry as an initial estimate for point cloud registration. In the first variant, we estimate the full 6-degree-of-freedom pose in 3D space, referred to as \threedkiss. In the second variant, the motion is constrained to a 2D plane, simplifying the estimation process. This is denoted as \twodkiss.

In addition to comparing our method with various \kiss variants, we further evaluate its performance by benchmarking it against two widely used state estimation methods in the robotics industry~\cite{macenski2023desks}. Specifically, we focus on integrating the most effective \kiss variant, \twodkiss, with the Fuse framework\cite{macenski2023desks} and the \textit{robot\_localization} package~\cite{moore2016ias}. The Fuse framework is a fixed-lag smoother that optimizes a pose graph by incorporating wheel odometry, LiDAR odometry, and kinematic constraints to estimate the robot's position. Meanwhile, \textit{robot\_localization} is an extended Kalman filter (EKF) that fuses sensor data to estimate the robot's state.

These two approaches involve computing LiDAR odometry and then fusing it with the wheel odometry. In contrast, our method directly processes both wheel odometry and LiDAR scans to produce the new pose in a single streamlined process. This approach generally uses less CPU and memory, reducing latency during live operation on the robot.

\subsubsection{Metrics}
\label{sec:metrics}
To evaluate the different methods, we use the average translation error from the \kitti benchmark, which calculates the average relative error over various trajectory lengths in percent. This metric is commonly used in autonomous driving, where the segment lengths for evaluation are considerably large for a wheeled mobile robot. To address this, we use smaller interval lengths of 1, 2, 5, 10, 20, 50, and 100\,m. Additionally, we report the root mean squared absolute translation error (ATE) after alignment as a measure of the global drift in the estimated trajectories.

\subsection{Qualitative Results in Large Indoor Warehouses}
\label{sec:qualitative}
The first experiment demonstrates the capability of our approach to correct the wheel odometry of a mobile robot while enforcing the kinematic constraints of the platform. We qualitatively assess the shape of the resulting trajectories by knowing that the robot moved in rectangular racks. For reasons of space, we show a portion of the Warehouse Small map, but the results are equally impressive for the other two sequences used for evaluation.

\begin{figure*}[t]
	\centering
	\def\svgwidth{0.9\linewidth}
	\begingroup%
  \makeatletter%
  \providecommand\color[2][]{%
    \errmessage{(Inkscape) Color is used for the text in Inkscape, but the package 'color.sty' is not loaded}%
    \renewcommand\color[2][]{}%
  }%
  \providecommand\transparent[1]{%
    \errmessage{(Inkscape) Transparency is used (non-zero) for the text in Inkscape, but the package 'transparent.sty' is not loaded}%
    \renewcommand\transparent[1]{}%
  }%
  \providecommand\rotatebox[2]{#2}%
  \newcommand*\fsize{\dimexpr\f@size pt\relax}%
  \newcommand*\lineheight[1]{\fontsize{\fsize}{#1\fsize}\selectfont}%
  \ifx\svgwidth\undefined%
    \setlength{\unitlength}{554.90386242bp}%
    \ifx\svgscale\undefined%
      \relax%
    \else%
      \setlength{\unitlength}{\unitlength * \real{\svgscale}}%
    \fi%
  \else%
    \setlength{\unitlength}{\svgwidth}%
  \fi%
  \global\let\svgwidth\undefined%
  \global\let\svgscale\undefined%
  \makeatother%
  \begin{picture}(1,0.58853224)%
    \lineheight{1}%
    \setlength\tabcolsep{0pt}%
    \put(0,0){\includegraphics[width=\unitlength,page=1]{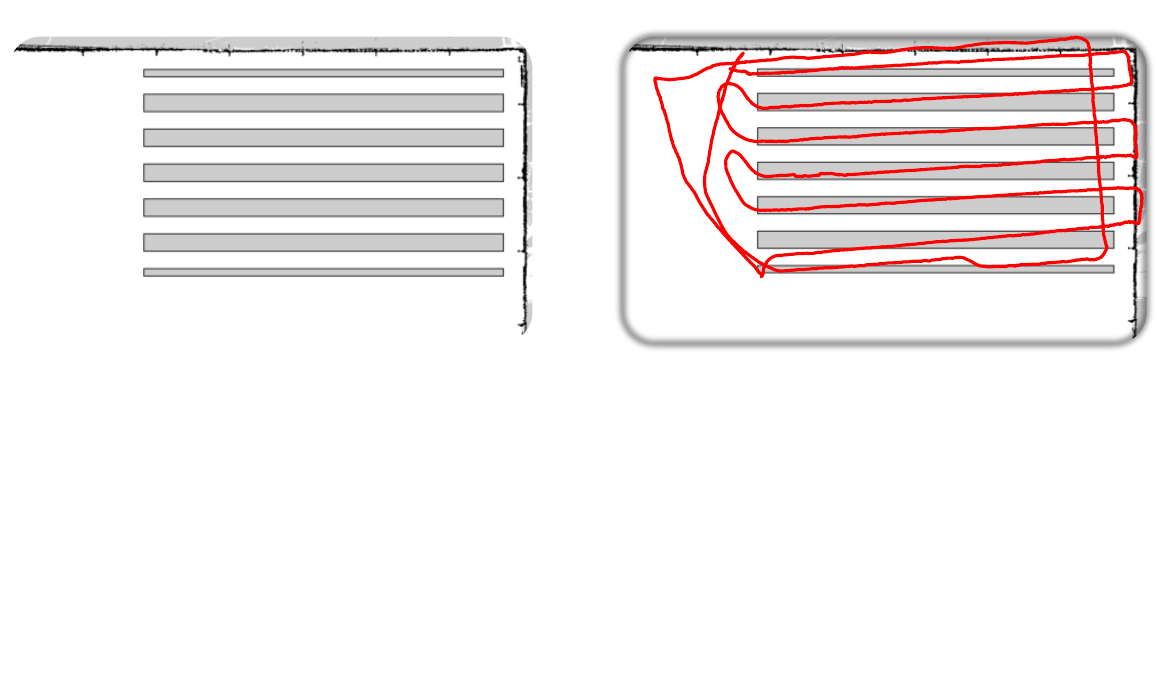}}%
    \put(0.1602026,0.30955602){\makebox(0,0)[lt]{\lineheight{1.25}\smash{\begin{tabular}[t]{l}Wheel Odometry\end{tabular}}}}%
    \put(0,0){\includegraphics[width=\unitlength,page=2]{dexory.pdf}}%
    \put(0.68057099,0.30550977){\makebox(0,0)[lt]{\lineheight{1.25}\smash{\begin{tabular}[t]{l}\twodkiss\end{tabular}}}}%
    \put(0,0){\includegraphics[width=\unitlength,page=3]{dexory.pdf}}%
    \put(0.70323542,0.02064718){\makebox(0,0)[lt]{\lineheight{1.25}\smash{\begin{tabular}[t]{l}\ours\end{tabular}}}}%
    \put(0,0){\includegraphics[width=\unitlength,page=4]{dexory.pdf}}%
    \put(0.15250903,0.02064718){\makebox(0,0)[lt]{\lineheight{1.25}\smash{\begin{tabular}[t]{l}\threedkiss\end{tabular}}}}%
  \end{picture}%
\endgroup%

	\caption{Qualitative comparison of odometry methods on Warehouse Small. Wheel odometry (top left) is smooth but drifts over time. \threedkiss (bottom left) lacks smoothness and accuracy. \twodkiss (top right) improves consistency but remains inaccurate. Our approach (bottom right) combines sensor data for smooth and accurate odometry estimation}
	\label{fig:dexory}
\end{figure*}

We first show on the top left the raw wheel odometry in~\figref{fig:dexory}, which follows the kinematic model of a differential drive but does drift over time. Next, we compare the result with two modified versions of KISS-ICP to account for the planar motion, and we observe that the racks become more visible, but the trajectories still bend. Note that the publicly available implementation of KISS-ICP will fail on this sequence due to the sparseness of the sensor~\cite{ress2024slam}. Besides that, the estimated trajectory is not smooth, which is unreasonable for a 12\,m tall robot that weighs approximately 500\,kg. Finally, using our proposed approach, the estimation is globally more accurate and still smooth. This is a direct consequence of enforcing the kinematic constraints in the optimization, which guarantees mathematically that the produced pose is always smooth and locally consistent.

\subsection{Quantitative Results}
\label{sec:quantitative}
In this section, we support our second claim, namely that our approach can estimate the odometry with a level of accuracy on par or better than state-of-the-art LiDAR odometry systems.~\tabref{tab:quantitative} showcases the results. First, one can see that our approach consistently achieves better results than the wheel odometry in terms of \rpe and \ate due to our proposed correction using the LiDAR data.

We also achieve better results than \kiss on all sequences, which is a state-of-the art LiDAR odometry system. Both relative and absolute errors of the KISS-ICP estimate are very large for the indoor warehouse environments, which shows how challenging these scenarios are due to the ambiguity of LiDAR-only measurements in featureless and repetitive corridors. Even when adding the wheel odometry as an initial guess, all the variants of KISS-ICP still underperform compared to other baselines. This can be explained by the fact that these approaches do not use a kinematic model for the optimization resulting in higher relative errors compared to our system.

The EKF and Fuse baselines perform well by combining LiDAR and wheel odometry, with Fuse often ranking second. However, our approach consistently outperforms both in terms of \ate and \rpe by directly processing the data and incorporating a kinematic model. Additionally, as noted earlier, our method is more computationally efficient since it optimizes wheel odometry and LiDAR corrections directly rather than fusing two different sources of odometry. For comparison, our system runs at 100\,Hz on a single-core CPU, while Fuse runs at approximately $10$\,Hz.

In the outdoor Palace sequence, our approach successfully corrects wheel odometry but performs slightly worse than KISS-ICP and its variants, see \tabref{tab:quantitative}. This sequence, recorded in a park with uneven terrain, challenges our assumption of a planar surface, as our kinematic model cannot account for factors like wheel slippage, rolling, or pitching. Despite these limitations, our method still provides reasonable odometry estimates and remains robust even in non-ideal, uneven conditions.

\subsection{Ablation on Regularization}
\label{sec:ablation}
Finally, our last experiment supports our third claim that our approach can dynamically adjust the influence of LiDAR scans and wheel odometry measurements, improving robustness and accuracy. To this end, we perform an ablation study comparing Kinematic-ICP with five different hand-picked values of~$\beta$ in~\eqref{eq:regularized_cost_simplified}, our adaptive regularization~\eqref{eq:adaption}, and without regularization. We selected Palace and Warehouse Small datasets for this experiment, as they are particularly challenging as demonstrated in~\secref{sec:quantitative}. In Palace, the robot moves on grass and rough terrain where the planarity assumption does not hold, while Warehouse Small is a sequence with mostly racks, resulting in a largely featureless environment.

\tabref{tab:ablation} presents the results of our ablation study. First, in these challenging scenarios, regularization is essential for accurate odometry estimation. This is particularly evident in the Warehouse Small dataset, a featureless environment where removing regularization causes the odometry accuracy to degrade by nearly an order of magnitude. Second, our adaptive regularization strategy consistently achieves strong results without requiring any fine-tuning of the~$\beta$ parameter. While some fixed~$\beta$ values yield the best performance, our method of computing it according to \eqref{eq:adaption} remains highly competitive. This highlights the robustness and flexibility of our approach, which performs well across different datasets without the need for manual adjustments, making it more flexible and widely applicable.

\begin{table}[t]
	\begin{tabularx}{\columnwidth}{l||Y|Y|Y|Y}
		\toprule
		                        & \multicolumn{2}{c|}{\textbf{Palace}} & \multicolumn{2}{c}{\textbf{Warehouse Small}}                                       \\
		\textbf{Regularization} & \rpe                                 & \ate                                         & \rpe             & \ate             \\
		\midrule
		Fixed $\beta=0.01$      & 2.39                                 & 1.70                                         & \B0.39           & \B 0.20          \\
		Fixed $\beta=0.1$       & \B1.79                               & 1.41                                         & 1.81             & 0.59             \\
		Fixed $\beta=1.0$       & 2.56                                 & 0.44                                         & 3.00             & 0.62             \\
		Fixed $\beta=10.0$      & 3.71                                 & \underline{0.33}                             & 3.03             & 1.75             \\
		Fixed $\beta=100.0$     & 3.99                                 & \B0.31                                       & 4.36             & 1.30             \\
		No Regularization       & 3.72                                 & 0.34                                         & 3.77             & 1.59             \\
		\ours                   & \underline{2.38}                     & 1.56                                         & \underline{0.53} & \underline{0.26} \\
		\bottomrule
	\end{tabularx}
	\caption{Ablation study on different regularization schemes for the optimization. We report the \rpe in [\%] and the \ate in [m]. The best results are in bold, and the second best are underlined.}
	\label{tab:ablation}
\end{table}

\vspace{0.3cm}
\section{Conclusion}
\label{sec:conclusion}
In this paper, we present Kinematic-ICP, a novel LiDAR odometry approach that explicitly incorporates the kinematic constraints of mobile robots into the classic point-to-point \ac*{icp} algorithm. Our approach exploits the knowledge of the unicycle motion model to estimate an odometry that is more consistent with the natural motion of a wheeled mobile platform. An adaptive regularization mechanism allows the system to adjust to degenerate conditions, ensuring robust performance even in scenarios challenging for traditional LiDAR odometry systems, such as feature-sparse environments. Our method provides accurate motion estimates by using kinematic constraints in the optimization to combine 3D LiDAR observations and wheel odometry.
We implemented and evaluated our approach in outdoor and large-scale warehouse environments, provided comparisons with existing state-of-the-art LiDAR odometry systems, and supported all the claims made in this paper. Our results demonstrate that Kinematic-ICP outperforms wheel odometry and existing LiDAR odometry techniques, delivering state-of-the-art accuracy.
Kinematic-ICP has been deployed in the Dexory fleet of robots operating in warehouses worldwide in production.

\section*{Acknowledgments}
We thank Gereon Tombrink for providing and supporting us with the total station reference system.

\bibliographystyle{plain_abbrv}

\bibliography{glorified,new}

\IfFileExists{./certificate/certificate.tex}{

\onecolumn

~\bigskip\bigskip\bigskip\bigskip %

	\section*{ \LARGE{Certificate of Reproducibility} }\vspace*{1cm}\Large{ 

	The authors of this publication declare that:}

	\vspace{5pt} 

	\begin{enumerate} 

		\setlength{\itemsep}{10pt} 

        \item The software related to this publication is distributed in the hope that it will be useful, support open research, and simplify the reproducability of the results but it comes without any warranty and without even the implied warranty of merchantability or fitness for a particular purpose.
  \item \textit{Tiziano Guadagnino and Benedikt Mersch} primarily developed the implementation related to this paper. This was done on  PopOS 22.04.

	\item \textit{Ignacio Vizzo} verified that the code can be executed on a machine that follows the software specification given in the Git repository available at: \\ 
\begin{center} 

  \url{https://github.com/PRBonn/kinematic-icp} \end{center}

 \item \textit{Ignacio Vizzo} verified that the experimental results presented in this publication can be reproduced using the implementation used at submission, which is labeled with a tag in the Git repository and can be retrieved using the command:\\

\begin{center} 

  \verb|git checkout v0.0.1| 

\end{center} 

\end{enumerate} 

\twocolumn

}{}
\end{document}